\pdfoutput=1

\documentclass[11pt]{article}

\usepackage[]{acl}

\usepackage{times}
\usepackage{latexsym}

\usepackage[T1]{fontenc}

\usepackage[utf8]{inputenc}
\usepackage{microtype}

\usepackage{makecell}
\usepackage{multirow}
\usepackage{diagbox}
\usepackage{pgfplots}
\usepackage{pgf-pie}
\usepackage{color}
\usepackage{mathtools}
\usepackage{amssymb}
\usepackage{graphicx}
\usepackage{subfigure} 
\usepackage{booktabs}

\usepackage{algorithm}
\usepackage{algorithmic}
\usepackage{hyperref}
\usepackage{xspace}
\newcommand{\method}{\textsc{SDM-Attack}\xspace}

\title{Modeling Adversarial Attack on Pre-trained Language Models as \\Sequential Decision Making}

\author{Xuanjie Fang\textsuperscript{\rm $1$}\thanks{\ \ Equal Contribution}, Sijie Cheng\textsuperscript{\rm $1,2,3,4$}\footnotemark[1], Yang Liu\textsuperscript{\rm $2,3,4,5$}, Wei Wang\textsuperscript{\rm $1$}\thanks{\ \ Corresponding Author} \\
$\textsuperscript{\rm $1$}$School of Computer Science, Fudan University, Shanghai, China \\
$\textsuperscript{\rm $2$}$Dept. of Comp. Sci. \& Tech., Institute for AI, Tsinghua University, Beijing, China \\
$\textsuperscript{\rm $3$}$Institute for AI Industry Research (AIR), Tsinghua University, Beijing, China \\
$\textsuperscript{\rm $4$}$Beijing National Research Center for Information Science and Technology, Beijing, China \\
$\textsuperscript{\rm $5$}$Shanghai Artificial Intelligence Laboratory, Shanghai, China \\
\small\texttt{\{xjfang20,sjcheng20,weiwang1\}@fudan.edu.cn}, \texttt{liuyang2011@tsinghua.edu.cn}
}

\begin{document}
\maketitle

\begin{abstract}
Pre-trained language models (PLMs) have been widely used to underpin various downstream tasks. However, the \textit{adversarial attack} task has found that PLMs are vulnerable to small perturbations. 
Mainstream methods adopt a detached two-stage framework to attack without considering the subsequent influence of substitution at each step.
In this paper, we formally model the adversarial attack task on PLMs as a \textit{sequential decision-making} problem, where the whole attack process is sequential with two decision-making problems, i.e., word finder and word substitution. 
Considering the attack process can only receive the final state without any direct intermediate signals, we propose to use reinforcement learning to find an appropriate sequential attack path to generate adversaries, named \method. Extensive experimental results show that \method achieves the highest attack success rate with a comparable modification rate and semantic similarity to attack fine-tuned BERT. Furthermore, our analyses demonstrate the generalization and transferability of \method.
The code is available at \url{https://github.com/fduxuan/SDM-Attack}.
\end{abstract}

\section{Introduction}

Nowadays, pre-trained language models (PLMs) have shown strong potential in various downstream tasks \citep{devlin2018bert, DBLP:journals/corr/abs-2005-14165}. 
However, a series of studies about \textit{adversarial attack} \citep{jin2020bert, li2020contextualized, li2020bert} have found that PLMs are vulnerable to some small perturbations based on the original inputs. The adversarial attack is essential to develop trustworthy and robust PLMs in Artificial Intelligence (AI) community \citep{thiebes2021trustworthy, marcus2020next}.


Despite the adversarial attack achieving success in both image and speech domains \citep{chakraborty2018adversarial,kurakin2018adversarial, carlini2018audio}, it is still far from perfect in the natural language processing (NLP) field due to the discrete nature of language \citep{studdert2005did, armstrong1995gesture}.
The main problem is to find an appropriate search algorithm that can make perturbations to mislead the victim models (i.e., PLMs) successfully~\citep{morris2020textattack,yoo2021towards}.
As mentioned in recent studies \citep{jin2020bert}, the challenges are preserving the following properties:
1) \textit{human prediction consistency}, misleading the PLMs while keeping human judges unchanged; 
2) \textit{semantic similarity}, keeping the semantics of the original inputs;
3) \textit{language fluency}, ensuring the correctness of grammar.

\begin{figure}[t]
    \centering
    \includegraphics[width=\linewidth]{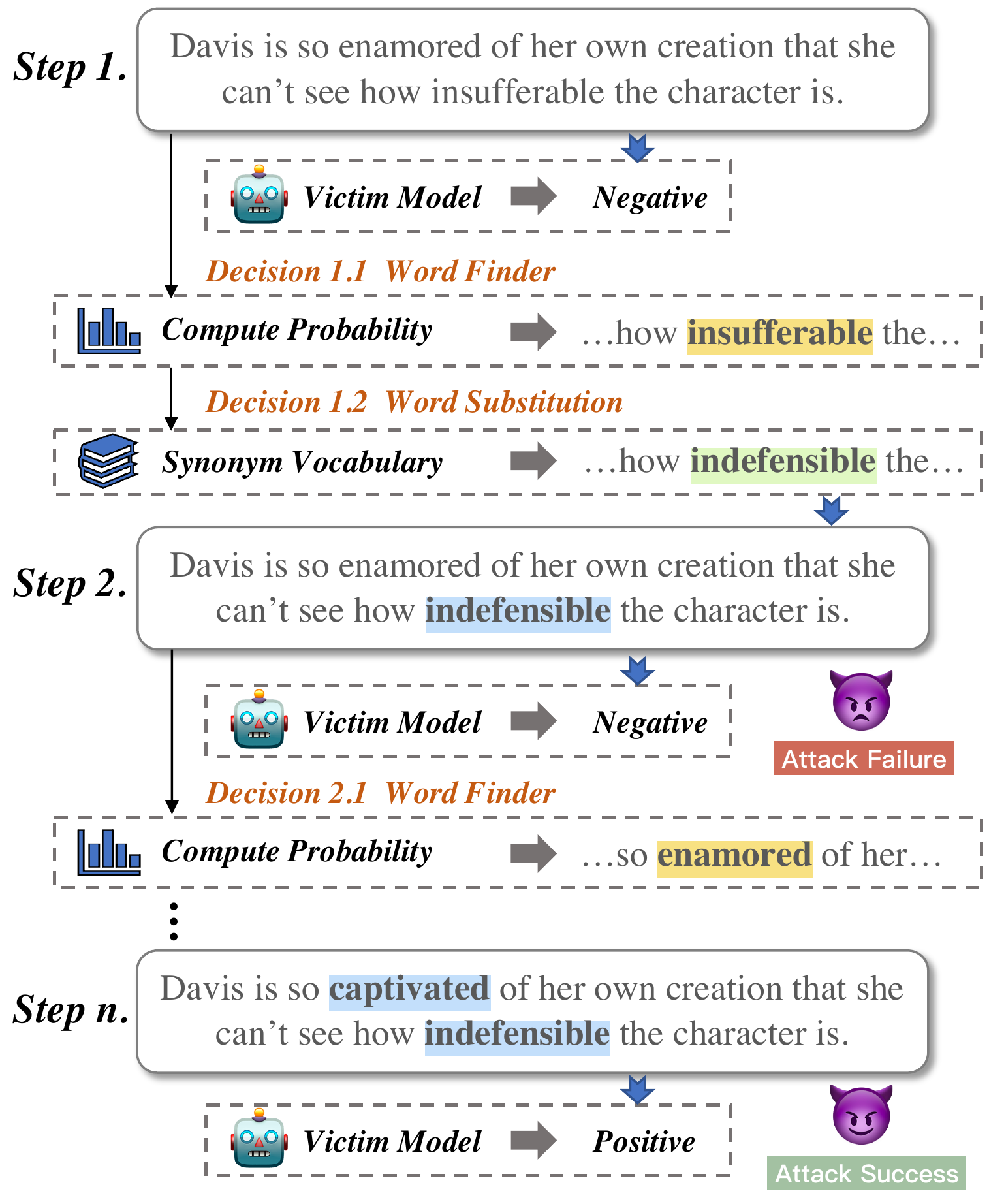}
    \caption{Illustrative example of modeling the adversarial attack into sequential decision making. The entire attack process is a sequence with two decision-making problems, i.e., word finder and substitution, until the adversary against the victim model is successful.}
    \label{fig:front}
\end{figure}

Mainstream solutions are typically a detached two-stage framework.
Specifically, they first rank the importance scores of all tokens according to the original input and then orderly substitute these tokens via heuristic rules.
Previous studies propose different strategies to rank the editing order of tokens, such as temporal-based  algorithm~\citep{gao2018black}, probability-weighted saliency~\citep{ren2019generating, li2020bert, li2020contextualized, jin2020bert}, and gradient-based ranking~\citep{yoo2021towards}.
However, these methods face two limitations.
On the one hand, they use a threshold to filter the unsatisfactory substitutions at last, but neglect to integrally consider the properties during computing importance scores.
On the other hand, their editing order only depends on the original input without considering the subsequent influence of substitution, as computing the importance score at each step is computationally burdensome in practice.

\begin{figure*}[t!]
    \centering
\includegraphics[width=\linewidth]{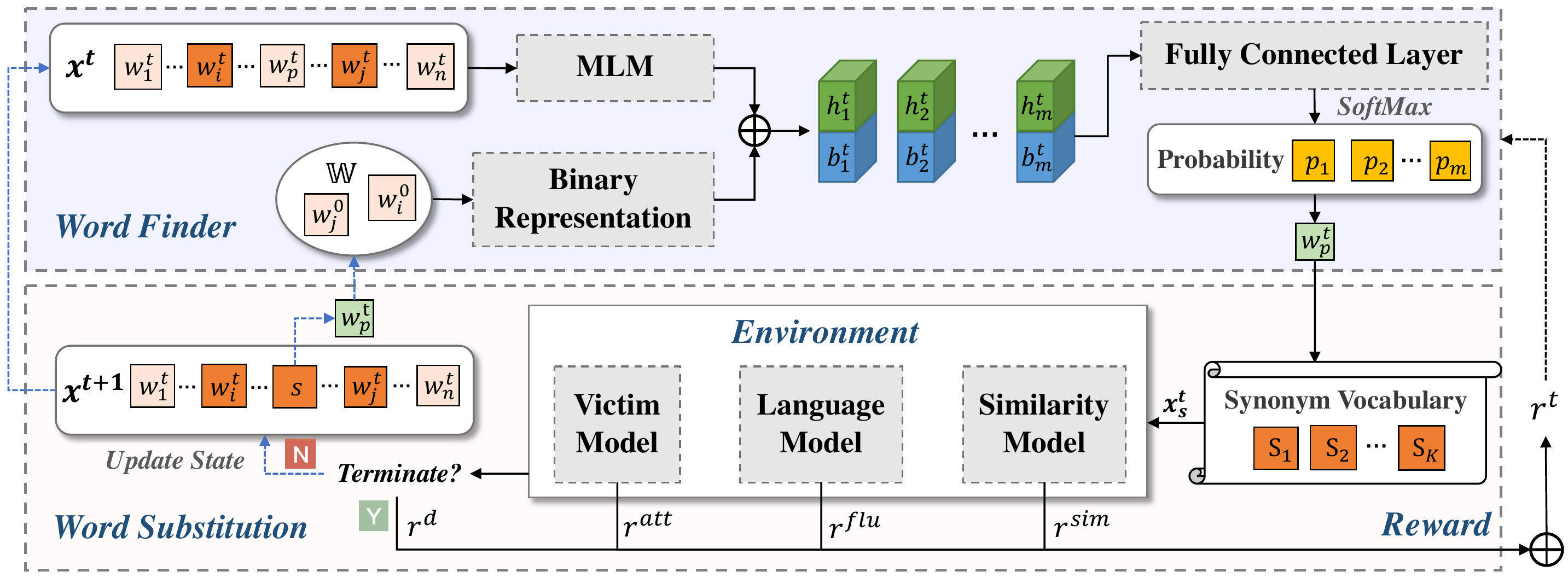}
    \caption{The framework of \method.}
    \label{fig:framework}
\end{figure*}
 
To solve the issues mentioned above, in this paper, we formally propose to transform the adversarial attack problem into a \textbf{sequential decision-making} task as shown in Figure \ref{fig:front}.
Rather than computing the importance scores all at once based on the original input, we regard the entire attack process as a sequence, where scores in the next step are influenced by the editing results in the current step.
Furthermore, there are two types of decision-making problems during each step in the attack sequential process: 1) \textit{word finder}, choosing the appropriate token to edit; 2) \textit{word substitution}, replacing the token with a suitable substitution.
Meanwhile, selecting edited tokens at each step should take the attack success rate and crucial properties, such as fluency, into account.

As a sequential decision-making task without a direct signal in each step, we naturally leverage reinforcement learning (RL) to find an appropriate sequential attack path to generate adversaries.
In this paper, we propose a model-agnostic method based on policy-based RL for modeling the adversarial attack into \underline{S}equential \underline{D}ecision \underline{M}aking, entitled \method. 
Given the victim model as the environment  with designed reward functions and the original input text as the initial state, the reinforced agent needs to decide on tokens to edit and synonyms to replace sequentially, until it attacks successfully.
The experimental results show that \method achieves the highest attack success rate with a comparable modification rate and semantic similarity to attack fine-tuned BERT against state-of-the-art baselines. 
Furthermore, we also demonstrate the effectiveness, generalizability, and transferability of \method in our analysis.

 

The main contributions of this work are summarized as the following:

\begin{itemize}
    \item To the best of our knowledge, we are the first to model the adversarial attack on PLMs into a sequential decision-making problem, where the whole attack process is sequential with two decision-making problems, i.e., word finder and word substitution.
    \item Considering the sequential attack process can receive the final state without any direct intermediate signals, we propose \method to use reinforcement learning to ask the agent to find an appropriate attack path based on our designed indirect reward signals yielded by the environment.


\end{itemize}

\section{Preliminaries}

As for NLP tasks, given a corpus of $N$ input texts ${\mathbb{X}} = \{\boldsymbol{x}_1, \boldsymbol{x}_2, \boldsymbol{x}_3,\cdots,\boldsymbol{x}_N\}$ and an output space ${Y} = \{y_1, y_2, y_3,\cdots,y_K\}$ containing $K$ labels, the language model $\mathrm{F}$ learns a mapping $f: \boldsymbol{x} \rightarrow y$ , which learns to classify each input sample $\boldsymbol{x} \in \mathbb{X}$ to the ground-truth label $y_{\text{gold}}\in \cal Y$:
\begin{equation}
 \mathrm{F}(\boldsymbol{x}) = \mathop{\arg\max}\limits_{y_i \in Y} P(y_i | \boldsymbol{x})
\end{equation}

The adversary of text $\boldsymbol{x} \in \mathbb{X}$ can be formulated as $\boldsymbol{x}_\text{adv} = \boldsymbol{x} + \epsilon$, where $ \epsilon$ is a slight perturbation to the input $\boldsymbol{x}$. The goal is to mislead the victim model $\mathrm{F}$ within a certain constraint $C(\boldsymbol{x}_\text{adv})$:
\begin{equation}
\begin{aligned}
&\mathrm{F}(\boldsymbol{x}_\text{adv}) = \mathop{\arg\max}\limits_{y_i \in Y}P(y_i|\boldsymbol{x}_\text{adv})\neq \mathrm{F}(\boldsymbol{x}),
\\
& \text{and}\ C(\boldsymbol{x}_\text{adv}, \boldsymbol{x}) \geq \lambda
\end{aligned}
\end{equation}
where $\lambda$ is the coefficient, and $C(\boldsymbol{x}_\text{adv}, \boldsymbol{x})$ usually calculates the semantic or syntactic similarity~\citep{cer2018universal,oliva2011symss} between the input $\boldsymbol{x}$ and its corresponding adversary $\boldsymbol{x}_\text{adv}$. 

Recently, the adversarial attack task has been framed as a combinatorial optimization problem. However, previous studies~\citep{gao2018black, ren2019generating, yoo2021towards} address this problem without considering the subsequent influence of substitution at each step, making attack far from the most effective.
In this paper, we formally define the adversarial attack as a sequential decision-making task, where the decisions in the next step are influenced by the results in the current step.



\section{Methodology}

In this section, we model the adversarial attack on PLMs problem as a sequential decision-making task as shown in Figure \ref{fig:front}, where the entire attack process is a sequence with two decision-making problems.
Considering the lack of direct signal in each step during the attack process, we propose a model-agnostic method, named \method, based on policy-based reinforcement learning. The illustration is shown in Figure \ref{fig:framework}.
During each step in the attack process, the reinforced agent needs to take two actions: 1) \textit{word finder}, choosing the appropriate token to edit; and 2) \textit{word substitution}, replacing the token with a suitable substitution. Through an attack sequence toward the input, we obtain its adversary until the attack is successful.


\subsection{Environment and Rewards} \label{sec:env}
We regard the victim models (i.e., PLMs) as the whole environment.
Intuitively, the agent needs to generate adversaries against the environment and achieve as high a reward as possible.
The $t$-step environment state is our intermediate generation $\boldsymbol{x}^{t} = [\boldsymbol{w}_1^{t}, \boldsymbol{w}_2^{t}, ..., \boldsymbol{w}_n^{t}]$ containing $n$ words, where the initial state $\boldsymbol{x}^{0}$ is the original input.

Considering the lack of direct signal in each step, our reward consists of a final discriminant signal $r_d$ to present the state of termination and an instant reward $r_t$ on every step.
As for the final signal $r_d$, once the model prediction of $t$-step state is different from the initial state, the environment will terminate this episode and yield a \textit{success} signal. 
However, if the model prediction does not change when all the tokens are replaced or the maximum number of steps is reached, a \textit{failure} signal will be given. Overall, the final signal $r_d$ is denoted as:
\begin{equation}
r_d = \begin{cases}
1, & \textit{success} \\
-1, & \textit{failure}
\end{cases}
\end{equation}

As for the instant reward $r_i$ for each step, we hope that the $t$-step state $\boldsymbol{x}^t$ can not only mislead the victim model but also ensure semantics similarity and fluency. Firstly, we design one instant reward to evaluate attack success rates:
\begin{equation}
r^{att}_t =\begin{cases}
r_d , \ \ \textit{terminated}\\
P(y_\text{gold}|\boldsymbol{x}^{t-1}) - P(y_\text{gold}|\boldsymbol{x}^t) ,\ \ \textit{survive}
\end{cases}
\label{equ:reward}
\end{equation}
where $r_d$ is the final reward if the current episode is terminated.
Secondly, we define a punishment by using an auto-regressive language model (LM) to measure fluency: 
\begin{equation}
r_t^{flu} =     \sum_i \frac{1}{|\boldsymbol{x}^t|}(\text{LM}(x_i | \boldsymbol{x}^t)  - \text{LM}(x_i | \boldsymbol{x}^{t-1}))
\end{equation}
where $\text{LM}(x_i|\boldsymbol{x}^t)$ is the cross-entropy loss of the token $x_i$ in sentence $\boldsymbol{x}^t$.
Thirdly, we also add semantic similarity constraints as another punishment:
\begin{equation}
r_t^{sim}  = \text{Sim}(\boldsymbol{x}, \boldsymbol{x^{t-1}}) - \text{Sim}(\boldsymbol{x}, \boldsymbol{x}^t)
\end{equation}
Finally, our overall instant reward $r_t$ is defined as:
\begin{equation}
r_t = \beta_1 r^{att}_t - \beta_2 r^{flu}_t -\beta_3 r^{sim}_t
\label{equ:instant_reward}
\end{equation}

\subsection{Decision Making} \label{sec:policy}

During each step in the whole attack process, there are two types of decision-making problems.
The first is choosing the appropriate token to edit, while the second is replacing the token with a suitable substitution. 
In RL, the agent needs to determine the decisions according to the yielded rewards.

\paragraph{Word Finder}
To find the appropriate token to edit, we first employ the masked language models (MLM) as an encoder to represent the state $\boldsymbol{x}^{t}$.
Due to the setup of the sub-word tokenizer in MLM, the encoder first converts $\boldsymbol{x}^{t}$ to a token sequence $\boldsymbol{x}^t_{token}=[\boldsymbol{o}_1^{t}, \boldsymbol{o}_2^{t}, ..., \boldsymbol{o}_{m}^{t}]$.
We reverse the conversion mapping $\phi: \boldsymbol{x}^t_{token} \rightarrow \boldsymbol{x}^{t}$ to recover tokens into words in need. Then we obtain the hidden states $\boldsymbol{h}^t=[\boldsymbol{h}^{t}_1, \boldsymbol{h}^{t}_2, ..., \boldsymbol{h}^{t}_m]$, where $\boldsymbol{h}_i^{t} \in {\Bbb R}^d$ is the hidden state of token $\boldsymbol{o}_i^{t}$ with $d$ dimensions. 

Furthermore, we maintain a word set $\mathbb{W}$ to restore the words of $\boldsymbol{x}$ that have been already modified as well as stop words and punctuation.
We then adopt a simple binary representation $\boldsymbol{b}^{t}$ according to the word set $\mathbb{W}$:
\begin{equation}
 \boldsymbol{b}_i^{t} = \begin{cases}
    {\bf 0} \in {\Bbb R}^d & \phi(\boldsymbol{o}_i^t) \in \mathbb{W} \\
    {\bf 1} \in {\Bbb R}^d & \phi(\boldsymbol{o}_i^t) \notin \mathbb{W}
    \end{cases}
\end{equation}

Then, we fuse both the hidden states $\boldsymbol{h}_i^t$ and the binary representation $\boldsymbol{b}_i^t$ to obtain the final representation $\boldsymbol{e}_i^t$ of the environment states:
\begin{equation}
\boldsymbol{e}_i^t = [\boldsymbol{h}_i^t; \boldsymbol{b}_i^t]
\end{equation}
where [; ] denotes the concatenation operation.

During the process of training, we first adopt a simple linear layer to obtain the probability and further normalize it into a distribution.
The probability distribution $p(\boldsymbol{o}^t_i|\boldsymbol{x}^t)$ of each token at $t$-step can be calculated as follows:
\begin{equation}
\begin{aligned}
p(\boldsymbol{o}^t_i|\boldsymbol{x}^t) = \mathrm{softmax}(W\cdot  \boldsymbol{e}^t_i + b)
\end{aligned}
\label{equ:sample}
\end{equation}where $W, b$ are the weight matrix and the bias vector, respectively. Then the agent samples the word $\boldsymbol{w}^t$ to substitute according to the distribution and ensures the sampled word is not in the word set $\mathbb{W}$.

During the evaluation, the agent will directly select the token with the maximum probability at each step, which is formulated as follows:
\begin{equation}
\boldsymbol{w}^t = \mathop{\arg\max}p(\boldsymbol{o}^t_i|\boldsymbol{x}^t), \phi(\boldsymbol{o}^t_i) \notin \mathbb{W}
\end{equation}

If the selected token $\boldsymbol{w}^t$ is a sub-word, we reverse the sub-word into a complete word via the conversion mapping $\phi$ as the newly selected word.

\paragraph{Word Substitution}
Following \citet{jin2020bert}, we adopt synonym substitution as our strategy after obtaining selected word $\boldsymbol{w}^t$ in $t$-step.
Firstly, we gather a synonym set $\mathbb{S}_{\boldsymbol{w}^t}$ for $\boldsymbol{w}^t$ that contains top-$k$ candidates from the external vocabulary, computing via cosine similarity ~\citep{mrkvsic2016counter}. 
Then, for each $\boldsymbol{s} \in \mathbb{S}_{\boldsymbol{w}^t}$, we replace $\boldsymbol{w}^t_p$ with $\boldsymbol{s}$ in the sentence $\boldsymbol{x}^{t}$ to get a substitution $\boldsymbol{x}^{t}_{\boldsymbol{s}} = [\boldsymbol{w}_1, ..., \boldsymbol{w}_{p-1}, \boldsymbol{s}, \boldsymbol{w}_{p+1}, ... \boldsymbol{w}_n]$.  
Finally, according to the instant reward $r_t$ in the Equation~\ref{equ:reward}, we select the substitution with the highest reward as the final adversaries $\boldsymbol{x}_\text{adv}^{t}$. Meanwhile, the environment states further updates as follows:
\begin{equation}
\begin{cases}
\boldsymbol{x}^{t+1} = \boldsymbol{x}^{t}_\text{adv} \\
\mathbb{W} = \mathbb{W} \cup \{\boldsymbol{w}^t_p\}
\end{cases}
\end{equation}


\subsection{Agent Training} 
The training target is to maximize the total return $G(\tau)$, which is an accumulated reward based on the instant reward $r_t$ , defined in Equation~\ref{equ:instant_reward}, with a discount factor $\gamma \in [0,1)$:
\begin{equation}
G(\tau) = \sum^T_{t=1} \gamma^t r_t
\end{equation}

The expected return of the decision trajectory, i.e., attack path, is defined as follows:
\begin{equation}
J(\theta) = \mathbb{E}[G(\tau)]
\end{equation}

Furthermore, we regard the agent as $\pi_\theta$ with parameters $\theta$ and the attack path as $\tau=[(a_1^f, a_1^s), \cdots,(a_T^f, a_T^s)]$, where $a_t^f$ and $a_t^s$ represent actions of \textit{word finder} and \textit{substitution} in $t$-th step, respectively.  The probability of this attack path is calculated as $\pi_\theta(\tau) = \prod_{t=1}^T \pi_\theta((a^f_t, a_t^s) | s_t)$,
where $\pi_\theta((a^f_t, a_t^s) | s_t) $ is the probability of actions in step $t$ based on current environment state $s_t$. Meanwhile, we consider $a_t^s$ a prior knowledge so that this probability can be simplified.
The gradient is calculated by REINFORCE algorithm \cite{kaelbling1996reinforcement}:
\begin{equation}
\nabla J(\theta) =\nabla  \mathbb{E} [\log \pi_\theta(\tau) \cdot G(\tau)]
\end{equation}
Detailed information of reinforce training is shown in appendix \ref{app:alg}. 


\begin{table*}[t!]
\centering
\small
\resizebox{\linewidth}{!}{
\begin{tabular}{llrrr|llrrr}
\toprule
\textbf{Dataset} & \textbf{Method} &\textbf{A-rate}$\uparrow$ & \textbf{Mod}$\downarrow$ & \textbf{Sim}$\uparrow$ &\textbf{Dataset} & \textbf{Method} & \textbf{A-rate }$\uparrow$& \textbf{Mod} $\downarrow$& \textbf{Sim}$\uparrow$\\
\midrule
\multirow{4}{*}{\textbf{Yelp}}& A2T & 88.3 & \textbf{8.1} & 0.68 & \multirow{4}{*}{\textbf{IMDB}} & A2T & 89.9&\underline{4.4}& \underline{0.79} \\
& TextFooler & \underline{90.5} & 9.0 & \underline{0.69}  & & TextFooler & \underline{88.7} & 7.6 & 0.76\\
& BERT-Attack & 89.8 & 12.4 & 0.66  & &BERT-Attack& 88.2 & 5.3 & 0.78\\

& \method & \textbf{95.8} & \underline{8.2} & \textbf{0.71} & &\method& \textbf{91.4} & \textbf{4.1} & \textbf{0.82} \\
\midrule
\multirow{4}{*}{\textbf{AG's News}} & A2T & 53.7 & \textbf{13.5} & \textbf{0.57} & \multirow{4}{*}{\textbf{MR}} & A2T &58.5 & \underline{12.6} & \underline{0.55} \\
& TextFooler & 66.2 & 18.4 & 0.52 && TextFooler& 80.5 & 15.8 & 0.50  \\
& BERT-Attack & \underline{74.6} & 15.6 & 0.52 & & BERT-Attack & \underline{83.2}&12.8 & 0.52 \\
& \method & \textbf{77.9} & \underline{15.3} & \underline{0.53} & & \method & \textbf{85.6}&\textbf{12.3}&\textbf{0.57}\\
\midrule
\multirow{4}{*}{\textbf{SNLI}}& A2T &70.8 & 17.2 & 0.35 &\multirow{4}{*}{\textbf{MNLI}} & A2T & 66.0 & 14.4& 0.45 \\
& TextFooler & \underline{84.3}&17.2& \underline{0.38} & & TextFooler & 76.5 & 15.0& 0.45 \\
& BERT-Attack & 81.9 & \underline{16.5} & \underline{0.38} & & BERT-Attack &\underline{78.1} & \underline{14.0} & \underline{0.46} \\

& \method &\textbf{ 85.5} & \textbf{15.9} & \textbf{0.43} & & \method & \textbf{78.7} & \textbf{13.8 }& \textbf{0.49} \\
\bottomrule
\end{tabular}
}
\caption{Automatic evaluation results of attack success rate (A-rate), modification rate (Mod), and semantic similarity (Sim). $\uparrow$ represents the higher the better and $\downarrow$ means the opposite. The results of MNLI dataset are the average performance of MNLI-matched and MNLI-mismatched. The best results are \textbf{bolded}, and the second-best
ones are \underline{underlined}.}
\label{tab:main}
\end{table*}

\section{Experiments}
\subsection{Experimental Setups}
\paragraph{Tasks and Datasets}
Following \citet{li2020bert, jin2020bert}, we evaluate the effectiveness of \method mainly on two standard NLP tasks, text classification and textual entailment. As for text classification, we use diverse datasets from different aspects, including news topic classification (AG's News;~\citealp{zhang2015character}), sentence-level sentiment analysis (MR;~\citealp{pang2005seeing}) and document-level sentiment analysis (IMDB\footnote{https://datasets.imdbws.com/} and Yelp Polarity;~\citealp{zhang2015character}). As for textual entailment, we use a dataset of sentence pairs (SNLI;~\citealp{bowman2015large}) and a dataset with multi-genre (MultiNLI;~\citealp{williams2017broad}). The statistics of datasets and more details can be found in Appendix \ref{app:datasets}. Following~\citet{jin2020bert, alzantot2018generating}, we attack 1k samples randomly selected from the test set of each task.

\paragraph{Baselines} We compare \method with recent state-of-the-art studies: 1) TextFooler~\citep{jin2020bert}: find important words via probability weighted word saliency and then apply substitution with counter-fitted word
embeddings.  2) BERT-Attack~\citep{li2020bert}: use mask-predict approach to generate adversaries. 
3) A2T \cite{yoo2021towards}: adopt faster search with gradient-based word importance ranking algorithm. 
We use open-source codes provided by the authors and TextAttack tools \cite{morris2020textattack} to implement these baselines. Furthermore, to ensure fairness in comparing baselines and \method, we apply constraints to all methods following  \citet{morris2020textattack} in Appendix \ref{app:constraint}.

\paragraph{Victim Models} We conduct the main experiments on a standard pre-trained language model BERT following \citep{jin2020bert, li2020bert}.
To detect the generalization of \method, we explore the effects on more typical models as discussed in Section~\ref{sec:models}.
All victim models are pre-trained from TextAttack \cite{morris2020textattack}.

\paragraph{Implementation Details}
We adopt BERT as the MLM model in word finder and  GPT-2~\cite{radford2019language} to measure fluency when computing rewards. To keep instant reward and punishment in a similar range, we set the hyper-parameters $\beta_1$ to be 1, $\beta_2$ to be 1 and $\beta_3$ to be 0.2.   Moreover, the discount factor $\gamma$ is set to be $0.9$ to achieve a trade-off between instant reward and long-term return. We set the episode number as $M=200$ and the learning rate as $\alpha = $ $3e^{-6}$  with Adam as the optimizer. In word substitution, the parameter $K$ of the synonyms number is $50$. Our experiments are conducted on a single NVIDIA 2080ti.

\paragraph{Automatic Evaluation Metrics} 
Following previous studies \citep{jin2020bert, morris2020textattack}, we use the following metrics as the evaluation criteria. 
1) Attack success rate (A-rate): the degraded performance after attacking target model.
2) Modification rate (Mod): the percentage of modified words comparing to original text.
3) Semantic similarity (Sim): the cosine similarity between the original text and its adversary, computing via the universal sentence encoder (USE; \citealp{cer2018universal}).

\paragraph{Manual Evaluation Metrics}
We further manually validate the quality of the adversaries from three challenging properties.
1) Human prediction consistency (Con): the rate of human judgement which is consistent with ground-truth label;
2) Language fluency (Flu): the fluency score of the sentence, measured on a Likert scale of 1 to 5 from ungrammatical to coherent \cite{gagnon2019salsa}; 
3) Semantic similarity ($\text{Sim}_\text{hum}$): the semantic consistency between each input-adversary pair, where 1 means \textit{unanimous}, 0.5 means \textit{ambiguous}, 0 means \textit{inconsistent}.

\subsection{Results}

\begin{table}[t!]
    \small
    \centering
    \begin{tabular}{llrrc}
    \toprule
         \textbf{Dataset} &  & \textbf{Con}$\uparrow$  & \textbf{Flu}$\uparrow$ & 
         $\textbf{Sim}_\text{hum}$ $\uparrow$\\
    \midrule
    \multirow{2}{*}{\textbf{IMDB}} & Original  & 0.95  & 4.5 & \multirow{2}{*}{0.95}\\
    & \method & 0.90 & 4.3 &  \\
    
     \midrule
    \multirow{2}{*}{\textbf{MNLI}} & Original  & 0.88  & 4.0 & \multirow{2}{*}{0.83}\\
    & \method & 0.79 & 3.7 &  \\
    
    \bottomrule
    \end{tabular}
    \caption{Manual evaluation results comparing the original input and generated adversary by \method of human prediction consistency (Con), language fluency (Flu), and semantic similarity ($\text{Sim}_\text{hum}$).}
    \label{tab:manual}
\end{table}

\paragraph{Automatic Evaluation}
As shown in Table \ref{tab:main}, \method consistently achieves the highest attack success rate to attack BERT in both text classification and textual entailment tasks, which indicates the effectiveness of \method.  
Furthermore, \method mostly obtains the best performance of modification and similarity metrics, except for AG's News, where \method achieves the second-best. 
For instance, our framework only perturbs 4.1\% of the words on the IMDB datasets, while the attack success rate is improved to 91.4\% with a semantic similarity of 0.82. 
Although A2T performs better in modification and similarity metrics in Yelp and AG's News, their attack success rate is always much lower than \method, even other baselines.
Because the modification and similarity metrics only consider the successful adversaries, we conjecture that A2T can only solve the inputs which are simpler to attack.
In general, our method can simultaneously satisfy the high attack success rate with a lower modification rate and higher similarity.
Furthermore, We find that the attack success rate on document-level datasets, i.e., Yelp and IMDB, are higher than the other sentence-level datasets, which indicates that it is easier to mislead models when the input text is longer. The possible reason is the victim model tends to use surface clues rather than understand them to make predictions when the context is long.

\paragraph{Manual evaluation}
In manual evaluation, we first randomly select 100 samples from successful adversaries in IMDB and MNLI datasets and then ask three crowd-workers to evaluate the quality of the original inputs and our generated adversaries.
The results are shown in Table \ref{tab:manual}.
As for the human prediction consistency, we regard the original inputs as a baseline.
Taking IMDB as an example, humans can correctly judge 95\% of the original inputs while they can maintain 90\% accuracy to our generated adversaries, which indicates \method can mislead the PLMs while keeping human judges unchanged.
The language fluency scores of adversaries are close to the original inputs, where the gap scores are within 0.3 on both datasets. 
Furthermore, the semantic similarity scores between the original inputs and our generated adversaries are 0.95 and 0.83 in IMDB and MNLI, respectively.
In general, \method can satisfy the challenging demand of preserving the three aforementioned properties. Detailed design of manual evaluation and more results are shown in appendix \ref{app:mannual}. 

\section{Analyses}

\subsection{Generalization}\label{sec:models}

\begin{table}[t!]
    \centering
    \small
    \begin{tabular}{llrrr}
    \toprule
    \textbf{Dataset} & \textbf{Model} & \textbf{A-rate}$\uparrow$ & \textbf{Mod}$\downarrow$ & \textbf{Sim}$\uparrow$\\
    \midrule
  & RoBERTa  & 84.4  & 13.9 & 0.52 \\
    
       \textbf{MR} & WordCNN & 72.1 & 10.3 & 0.48 \\
     & WordLSTM & 80.7 & 8.9 & 0.56 \\
    \midrule
   & RoBERTa  &88.3 & 8.3 & 0.70\\
    
       \textbf{IMDB} & WordCNN & 89.2  & 3.3 & 0.85 \\
     & WordLSTM & 89.8 & 5.4 &  0.75\\
     \midrule
    \multirow{2}{*}{\textbf{SNLI}}  & InferSent & 78.7 &17.0 & 0.42 \\
    & ESIM & 79.0 &17.2  & 0.41\\
     \bottomrule
   \end{tabular}
    \caption{Attack results against other models.}
    \label{tab:other}
\end{table}

\begin{table}[t!]
\centering
\small
\resizebox{\linewidth}{!}{
\begin{tabular}{llccc}
\toprule
\textbf{Dataset} & \textbf{Method} & \textbf{A-rate}$\uparrow$ & \textbf{Mod}$\downarrow$ & \textbf{Sim}$\uparrow$\\
\midrule
\multirow{2}{*}{\textbf{AG's News}} & BERT-Attack & 74.6 & 15.6 & 0.52 \\
&  \method-mlm  & 76.2 & 15.0 & 0.51 \\
\midrule
\multirow{2}{*}{\textbf{MR}} &BERT-Attack& 83.2&12.8&0.52 \\
&  \method-mlm & 84.3 & 11.5 & 0.53 \\
\bottomrule
\end{tabular}
}
\caption{Attack results of different substitution strategies, where \method-mlm is replaced with the same strategy of word finder as BERT-Attack.} 
\label{tabel:generalization}
\end{table}

We detect the generalization of \method in two aspects, 1) attack more language models and 2) adapt to more substitution strategies.
Firstly, we apply \method to attack extensive victim models, such as traditional language models (e.g., WordCNN) and other state-of-the-art PLMs (e.g., RoBERTa; \citealp{liu2019roberta}).
The results of text classification tasks in table \ref{tab:other} show that \method not only has better attack effects against WordCNN and WordLSTM, but also misleads RoBERTa, which is a more robust model.
For example, on the IMDB datasets, the attack success rate is up to 89.2\% against WordCNN with a modification rate of only about 3.3\% and a high semantic similarity of 0.85. 
As for the textual entailment task, \method can also achieve remarkable attack success rates against InferSent and ESIM.

Secondly, although we directly adopt the word substitution strategy in Textfooler, this strategy can actually be replaced by other methods.
To demonstrate this assumption, we further replace our word substitution strategy with the mask-fill way in BERT-attack, named \method-mlm.
As shown in Table \ref{tabel:generalization}, \method-mlm completely beat BERT-Attack, indicating the part of word substitution of \method has generalization ability to extend to different types of strategies and archives high performance. More results are displayed in appendix \ref{app:mannual}.

\subsection{Efficiency}

\begin{figure}[t!]
\includegraphics[width=\linewidth]{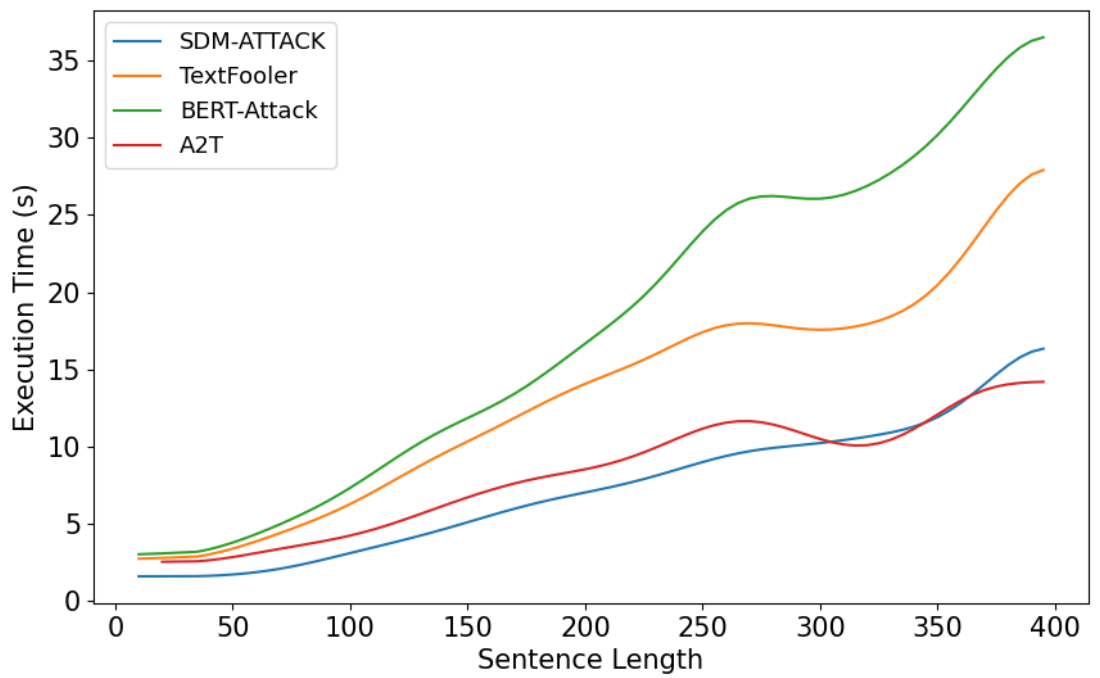}
\caption{The time cost according to varying sentence lengths in the IMDB dataset, smoothed with Gaussian function where kernel size is 5. } 
\label{effect}
\end{figure}

In this section, we probe the efficiency according to varying sentence lengths in the IMDB dataset as shown in Figure \ref{effect}. 
The time cost of \method is surprisingly mostly better than A2T, which mainly targets obtaining cheaper computation costs with lower attack success rates in Table \ref{tab:main}. 
Meanwhile, \method can obviously beat BERT-attack and TextFooler, which need to conduct a model forward process for each token.
Furthermore, with the increase of sentence lengths, \method and A2T maintain a stable time cost, while the time cost of BERT-attack and TextFooler is exploding. These phenomena show the efficiency advantage of \method, especially in dealing with long texts.


\subsection{Transferability}

We evaluate the transferability of \method to detect whether the \method trained on one dataset can perform well on other datasets.
We conduct experiments on a series of text classification tasks and use the randomly initialized BERT as a baseline.
As shown in Table \ref{tab:transfer2}, \method has high transferability scores across different datasets, which are consistently higher than random.
In detail, the performances among Yelp, IMDB and MR, which all belong to sentiment analysis, are higher than AG's News.
Moreover, IMDB and MR are corpora about movies where \method tends to learn a general attack strategy in this field and can transfer well to each other.


    
    
     


\subsection{Adversarial Training}

\begin{table}[t!]
    \centering
    \small
    \begin{tabular}{ccccc}
    \toprule
         & \textbf{Yelp} & \textbf{IMDB} & \textbf{MR} & \textbf{AG's News} \\
    \midrule
     \textbf{Yelp}  & \textbf{87.6} & 85.8 & 40.5 & 43.6 \\
    
    \textbf{IMDB} & 82.9  & \textbf{89.3} & 51.4 & 43.4 \\
    \textbf{MR} & 81.8 & 88.2 &  \textbf{66.5} & 39.6 \\
    \textbf{AG's News} & 62.4 & 59.2 & 29.9 &  \textbf{53.2}\\
    \textbf{Random} & 58.9 & 56.1  & 27.8 & 38.3\\ 
     \bottomrule
   \end{tabular}
    \caption{Transferability evaluation of \method generator on text classification task against BERT. Row $i$ and column $j$ is the attack success rate of \method trained on dataset $i$  evaluated on dataset $j$.}
    \label{tab:transfer2}
\end{table}

\begin{table}[t!]
    \small
    \centering
    \begin{tabular}{lrrrr}
    \toprule
    \textbf{Dataset} & \textbf{Acc}$\uparrow$  & \textbf{A-rate}$\uparrow$ & \textbf{Mod}$\downarrow$ & \textbf{Sim}$\uparrow$ \\
      \midrule
     \textbf{Yelp} & 97.4  & 95.8 & 8.2 &0.71\\
     +Adv Train & 97.0 & 82.5 & 13.5 &0.63  \\ 
     \midrule
     \textbf{IMDB} &91.6 &91.4 & 4.1 & 0.82 \\
     +Adv Train  & 90.5 & 79.2 & 8.5 & 0.74 \\
    \midrule
     \textbf{SNLI} &89.1 & 85.5 & 15.9 & 0.43 \\
      +Adv Train & 88.2 & 78.6 & 17.1 & 0.42\\
     \bottomrule
   \end{tabular}
    \caption{The results of comparing the original training with adversarial training with our generated adversaries. More results can be found in Appendix \ref{app:training}.}
    \label{tab:adversarial_training}
\end{table}

\begin{table*}[t!]
\centering
\small
\begin{tabular}{lp{8cm}cccc}
\toprule 
\textbf{Method}  & \textbf{Text} (MR; Negative)& \textbf{Result} & \textbf{Mod}$\downarrow$ & \textbf{Sim}$\uparrow$ & \textbf{Flu}$\uparrow$ \\
\midrule
\multirow{2}{*}{Original} & Davis is so enamored of her own creation that she can not see how insufferable the character is. & \multirow{2}{*}{-}& \multirow{2}{*}{-}  & \multirow{2}{*}{-} & \multirow{2}{*}{5} \\
\midrule
 \multirow{2}{*}{A2T} & Davis is so enamored of her own \textcolor{blue}{institution} that she can not \textcolor{blue}{ behold} how \textcolor{blue}{unforgivable} the \textcolor{blue}{hallmark} is. &\multirow{2}{*}{\textit{Failure}}  & \multirow{2}{*}{22.2}& \multirow{2}{*}{0.16} & \multirow{2}{*}{3} \\
\midrule
 \multirow{2}{*}{TextFooler} & Davis is \textcolor{blue}{well} enamored of her own \textcolor{blue}{infancy} that she \textcolor{blue}{could}  not \textcolor{blue}{admire} how \textcolor{blue}{infernal} the \textcolor{blue}{idiosyncrasies} is. &\multirow{2}{*}{\textit{Success}} &\multirow{2}{*}{33.3} & \multirow{2}{*}{0.23} & \multirow{2}{*}{3} \\
\midrule
 \multirow{2}{*}{BERT-Attack} & Davis is \textcolor{blue}{often} \textcolor{blue}{enamoted} of her own \textcolor{blue}{generation} that she can not see how \textcolor{blue}{insuffoure} the \textcolor{blue}{queen} is. &\multirow{2}{*}{\textit{Failure}}& \multirow{2}{*}{27.8} & \multirow{2}{*}{0.09} & \multirow{2}{*}{2}\\
\midrule
\multirow{2}{*}{\method}  & Davis is so \textcolor{blue}{captivated} of her own creation that she can't see how \textcolor{blue}{indefensible} the character is. & \multirow{2}{*}{\textit{Success}}& \multirow{2}{*}{11.1} & \multirow{2}{*}{0.57} & \multirow{2}{*}{5} \\
\bottomrule
\end{tabular}
\caption{Adversaries generated by \method and baselines in MR dataset. The replaced words are highlighted in \textcolor{blue}{blue}. \textit{Failure} indicates the adversary fails to attack the victim model and \textit{success} means the opposite.}
\label{tab:case_study}
\end{table*}

We further investigate to improve the robustness of victim models via adversarial training.
Specifically, we fine-tune the victim model with both original training datasets and our generated adversaries, and evaluate it on the same test set.
As shown in Table \ref{tab:adversarial_training}, compared to the results with the original training datasets, adversarial training with our generated adversaries can maintain close accuracy, while improving performance on attack success rates, modification rates, and semantic similarity.
The victim models with adversarial training are more difficult to attack, which indicates that our generated adversaries have the potential to serve as supplementary corpora to enhance the robustness of victim models.

\subsection{Case Study}
Table \ref{tab:case_study} shows adversaries produced by \method and the baselines. Overall, the performance of \method is significantly better than other methods. For this sample from the MR dataset, only TextFooler and \method successfully mislead the victim model, i.e., changing the prediction from \textit{negative} to \textit{positive}. However, TextFooler modifies twice as many words as \method, demonstrating our work has found a more suitable modification path. Adversaries generated by A2T and BERT-Attack are failed samples due to the low semantic similarity. BERT-Attack even generates an invalid word ``\textit{enamoted}" due to its sub-word combination algorithm.  
We also ask crowd-workers to give a fluency evaluation. Results show \method obtains the highest score of 5 as  the original sentence, while other adversaries are considered difficult to understand, indicating \method can generate more natural sentences.


\section{Related Work}
Adversarial attack has been well-studied in image and speech domains~\citep{szegedy2013intriguing,chakraborty2018adversarial,kurakin2018adversarial, carlini2018audio}.
However, due to the discrete nature of language, the adversarial attack against pre-trained language models is much more difficult.
Earlier works mainly focus on designing heuristic rules to generate adversaries, including swapping words~\cite{wei2019eda}, transforming syntactic structure~\cite{coulombe2018text}, and paraphrasing by back-translation \cite{ribeiro2018semantically, xie2020unsupervised}.
However, these rule-based methods are label-intensive and difficult to scale.

Recently, adversarial attack in NLP is framed as a combinatorial optimization problem.
Mainstream studies design a series of search algorithms with two detached stages
In the first stage, they iteratively search for modification positions, including saliency-based ranking~\citep{liang2017deep,ren2019generating,jin2020bert,garg2020bae}, gradient-based descent algorithm~\citep{sato2018interpretable,yoo2021towards}, and temporal-based  searcher~\citep{gao2018black}.
In the second stage, a series of studies designs different substitution strategies, including dictionary method~\citep{ren2019generating}, word embeddings~\cite{kuleshov2018adversarial,jin2020bert} or language models~\cite{li2020bert,garg2020bae,li2020contextualized}.
In this paper, we formally propose to define the adversarial attack task as a sequential decision-making problem, further considering that scores in the next step are influenced by the editing results in the current step.

The other line of recent studies is sampling-based methods. \citet{alzantot2018generating} and \citet{wang1909natural} apply genetic-based algorithm, \citet{zang2019word} propose a particle swarm optimization-based method,  and \citet{guo2021gradient} generate adversaries via distribution approximate sampling. 
However, their execution time is much more expensive due to the properties of sampling, so it is unlikely to generate large-scale adversarial samples.
In addition, \citet{zou2019reinforced} conducts reinforcement learning on attacking the neural machine translation task, but their search path is fixed from left to right.
In this paper, \method can determine any search order to find the appropriate attack path.


\section{Conclusion}
In this paper, we formally define the adversarial attack task as a sequential decision-making problem, considering the entire attack process as sequence with two types of decision-making problems, i.e., word finder and substitution.
To solve this problem without any direct signals of intermediate steps, we propose to use policy-based RL to find an appropriate attack path, entitled \method.
Our experimental results show that \method achieves the highest attack success rate.
In this paper, we use our designed rewards as instant signals to solve these two decision-making problems approximately. We will further try to adopt hierarchical RL to optimize the solution.

\section{Limitations}
We define the adversarial attack task as a sequential decision-making problem and apply policy-based reinforcement learning to model it. This work must follow this assumption: the decision process conforms to Markov decision process (MDP) that the conditional probability distribution of the future state depends only on the current state. Meanwhile, reinforcement learning training requires additional time costs and the results may be unstable. 

We only conduct the experiments on two NLP tasks with six selected datasets, which are all English corpus.  Furthermore, our experimental results are mainly for BERT, with RoBERTa supplemented in the analysis. Thus, we lack the evaluation of other novel pre-trained language models, such as ELECTRA \cite{clark2020electra} and XLNET \cite{yang2019xlnet}. Therefore, our work lacks multi-task, multi-model and multilingual verification in terms of generalization and transferability.

\section{Ethics Statement}
We declare that this article is in accordance with the ethical standards of \textit{ACL Code of Ethics}. Any third party tools used in this work are licensed from their authors. All crowd-workers participating in the experiments are paid according to the local hourly wages.

\section{Acknowledgment}
We would like to thank anonymous reviewers for their insightful and constructive feedback. We appreciate Peng Li and Shuo Wang for their valuable discussions. We thank Qianlin Liu, Yanqi Jiang and Yiwen Xu for the crowdsourced work.
This work is supported by the National Key R\&D Program of China (2022ZD0160502) and the National Natural Science Foundation of China (No. 61925601, 62276152, 62236011).

\bibliography{anthology,custom}
\bibliographystyle{acl_natbib}

\appendix

\clearpage

\section{Datasets}
\label{app:datasets}
We conduct experiments on the following datasets of two NLP tasks and detailed statistics are displayed in Table \ref{tab:overall_datasets}: 
\begin{itemize}
\item \textbf{Text Classification}: (1) Yelp \citep{zhang2015character}: A dataset for binary sentiment classification on reviews, constructed by considering stars 1 and 2 negative, and 3 and 4 positive. (2) IMDB: A document-level movie review dataset for binary sentiment analysis.  (3) MR \cite{pang2005seeing}: A sentence-level binary classification dataset collected from Rotten Tomatoes movie reviews. (4) AG's News \cite{zhang2015character}:  A collection of news articles. There are four topics in this dataset: World, Sports, Business, and Science/Technology.
\end{itemize}
\begin{itemize}
\item \textbf{Textual Entailment}: (1) SNLI \citep{bowman2015large}:  A dataset of human-written English sentence pairs and manually annotated labels of entailment, neutral and contradiction. (2) MNLI \citep{williams2017broad}: Another crowd-sourced collection of sentence pairs labeled with textual entailment information. Compare to SNLI, it includes more complex sentences, e.g, enres of spoken and written text.
\end{itemize}

 \section{Training Algorithm}
 \label{app:alg}

The training process is shown in Algorithm \ref{train}. Since $a_t^f$ is chosen through a probability distribution, the agent is encouraged to explore more possible paths. 
The instant reward $r_t$ is obtained from environment after performing both two actions actions.
Once the termination signal is raised, the environment will terminate this current episode and update the agent's parameters via a policy gradient approach.
The expected return of decision trajectory is defined as follows:
\begin{equation}
J(\theta) = \mathbb{E}[G(\tau)]
\end{equation}Thus the gradient is calculated by REINFORCE algorithm \cite{kaelbling1996reinforcement}:
\begin{equation}
\nabla J(\theta) =\nabla  \mathbb{E} [\log \pi_\theta(\tau) \cdot G(\tau)]
\end{equation}
Then the expectation over the whole  sequence is approximated by Monte Carlo simulations and can be expressed as follows:
\begin{equation}
\nabla J(\theta) = \frac{1}{M} \sum^M_{m=1} \nabla \log \pi_\theta(\tau^{(m)})G(\tau^{(m)})
\end{equation} where  $[\tau^{(1)}, \tau^{(2)}, ..., \tau^{(M)}]$ are $M$ samples of trajectories.
The discount factor $\gamma$  enables both long-term and immediate effects to be taken into account and trajectories with shorter lengths are encouraged. 

We randomly select 2500 items from the training corpus for training the agent of each dataset. The average convergence time is approximately between 2-16 hours, related to the length of the input. When attacking large batches of samples, the impact of training cost is negligible compared to the cumulative attack time cost. During training, We adopt random strategies and short-sighted strategies in the initial stage for early exploration and to obtain better seeds. 
\begin{table}
\small
\begin{tabular}{lrrrr}
\toprule
\textbf{Dataset} & \textbf{Train} & \textbf{Test} & \textbf{Avg Len} & \textbf{Classes} \\
\midrule
Yelp &  560k & 38k & 152&2  \\
IMDB & 25k & 25k & 215&2  \\
AG's News& 120k & 7.6k & 73&4\\
MR & 9k & 1k & 20&2\\
\midrule
SNLI & 570k & 3k & 8 & 3\\
MNLI & 433k & 10k & 11& 3\\
\bottomrule
\end{tabular}
\caption{Overall statistics of datasets.}
\label{tab:overall_datasets}
\end{table}

\begin{algorithm}[t!]
    \caption{Reinforce Training}
    \begin{algorithmic}[1]
    	\STATE Initialization: agent $\pi_\theta$ with parameters $\theta$, episode number $M$
        		\FOR  {$i \gets 1$ to M}
                		\STATE initialize $t \gets 1$
                        \WHILE {not receive termination signal} 
                        		\STATE get environment state $s_t$
                                \STATE compute  $ \pi_\theta((a^f_t, a^s_t)|s_t)\backsim\pi_\theta(a_t^f|s_t)$
                                \STATE sample $a_t^f$ based on probability
                                \STATE select $a^s_t$ from prior knowledge
                                \STATE compute reward $r_t$ 
                                \STATE update $t \gets t+1$
                        \ENDWHILE
                        \STATE initialize $G(\tau)$ ← 0
                        \FOR { $j   \gets T $ to 1 }
                      		\STATE $G(\tau) \gets \gamma G(\tau) + r_j$  
                              \STATE accumulate $J_j(\theta)$
                        \ENDFOR
                        \STATE update $\theta \gets \theta + \alpha \nabla J(\theta)$
                \ENDFOR
    \end{algorithmic}
    \label{train}
\end{algorithm}

\section{Implementation Constraint}
\label{app:constraint}
In order to make the comparison fairer, we set the following constraints for \method as well as all baselines: (1) \textbf{Max modification rate}: To better maintain semantic consistency, we only keep adversarial samples with less than 40\% of the words to be perturbed.  (2) \textbf{Part-of-speech (POS)}:  To generate grammatical and fluent sentences, we use NLTK tools\footnote{https://www.nltk.org/} to filter candidates that have a different POS from the target word. This constraint is not employed on BERT-Attack. (3) \textbf{Stop words preservation}: the modification of stop words is disallowed and this constraint helps avoid grammatical errors. (4) \textbf{Word embedding distance}: For Textfooler, A2T and \method, we only keep candidates with word embedding cosine similarity higher than 0.5 from synonyms dictionaries \cite{mrkvsic2016counter}. For \textit{mask-fill} methods,  following BERT-Attack, we filter out antonyms \cite{li2020bert} via the same synonym dictionaries for sentiment classification tasks and textual entailment tasks.


\section{Tuning with Adversaries}
\label{app:training}

Table \ref{tab:training} displays adversarial training results of all datasets. Overall, after fine-turned with both original training datasets and adversaries, victim model is more difficult to attack. Compared to original results, accuracy of all datasets is barely affected, while attack success rate meets an obvious decline. Meanwhile, attacking model with adversarial training leads to higher modification rate, further demonstrating adversarial training may help improve robustness of victim models.
\begin{table}[t!]
    \small
    \centering
    \begin{tabular}{lrrrr}
    \toprule
    \textbf{Dataset} & \textbf{Acc}$\uparrow$  & \textbf{A-rate}$\uparrow$ & \textbf{Mod}$\downarrow$ & \textbf{Sim}$\uparrow$ \\
      \midrule
     \textbf{Yelp} & 97.4  & 95.8 & 8.2 &0.71\\
     +Adv Train & 97.0 & 82.5 & 13.5 &0.63  \\ 
     \midrule
     \textbf{IMDB} &91.6 &91.4 & 4.1 & 0.82 \\
     +Adv Train  & 90.5 & 79.2 & 8.5 & 0.74 \\
    \midrule
    \textbf{AG's News}  & 94.6 & 77.9 & 15.3 & 0.53 \\
    +Adv Train & 91.8 & 50.6 & 23.3 & 0.50\\
    \midrule
    \textbf{MR} & 96.9 & 85.6  &  12.3 & 0.57\\
     +Adv Train & 92.4 & 72.0 & 16.7 & 0.57 \\
     \midrule
     \textbf{SNLI} &89.1 & 85.5 & 15.9 & 0.43 \\
      +Adv Train & 88.2 & 78.6 & 17.1 & 0.42\\
      \midrule
     \textbf{MNLI} &84.5 & 78.7 & 13.8 & 0.49 \\
      +Adv Train & 76.8 & 58.6 & 15.2 & 0.49\\
     \bottomrule
   \end{tabular}
    \caption{Adversarial training results.}
    \label{tab:training}
\end{table}

\section{Supplementary Results}
\label{app:mannual}
At the beginning of manual evaluation, we provided some data to allow crowdsourcing workers to unify the evaluation standards. We also remove the data with large differences when calculating the average value to ensure the reliability and accuracy of the evaluation results. More manual evaluation results are shown in Table \ref{tab:manual_all}.

\begin{table}[t!]
    \small
    \centering
    \begin{tabular}{llrrc}
    \toprule
         \textbf{Dataset} &  & \textbf{Con}$\uparrow$  & \textbf{Flu}$\uparrow$ & 
         $\textbf{Sim}_\text{hum}$ $\uparrow$\\

    \midrule
    \multirow{3}{*}{\textbf{IMDB}} & Original  & 0.95  & 4.5 & \\
    & TextFooler & 0.84  & 4.0 & 0.88\\
    & Bert-Attack & 0.83  & 4.2 & 0.90\\
    & \method & 0.90 & 4.3 & 0.95 \\
    
     \midrule
    \multirow{2}{*}{\textbf{MNLI}} & Original  & 0.88  & 4.0 & \\
     & TextFooler & 0.77  & 3.5 & 0.80\\
     & Bert-Attack & 0.77  & 3.6 & 0.81\\
    & \method & 0.79 & 3.7 & 0.83 \\
    
    \bottomrule
    \end{tabular}
    \caption{Manual evaluation results comparing the original input and generated adversary by attack method of human prediction consistency (Con), language fluency (Flu), and semantic similarity ($\text{Sim}_\text{hum}$).}
    \label{tab:manual_all}
\end{table}

Table \ref{tabel:generalization_all} displays the generalization ability of \method with mask-fill strategy. However, the improvement effect is not particularly obvious. The mask-fill method makes the current candidate synonyms also affected by the sequence states. Compared to a fixed synonym dictionary, it has a larger prior knowledge and changing action space, which makes it harder to train the agent. Only increasing the size of the training corpus is not very effective. We will try adopting hierarchical RL to further solve this problem in the future.

\begin{table}[t!]
\centering
\small
\resizebox{\linewidth}{!}{
\begin{tabular}{llccc}
\toprule
\textbf{Dataset} & \textbf{Method} & \textbf{A-rate}$\uparrow$ & \textbf{Mod}$\downarrow$ & \textbf{Sim}$\uparrow$\\
\midrule
\multirow{2}{*}{\textbf{Yelp}} & BERT-Attack & 89.8 & 12.4 & 0.66 \\
&  \method-mlm  & 90.0 & 10.6 & 0.65 \\
\midrule
\multirow{2}{*}{\textbf{IMDB}} & BERT-Attack & 88.2 & 5.3 & 0.78 \\
&  \method-mlm  & 88.5 & 5.1 & 0.78 \\
\midrule
\multirow{2}{*}{\textbf{AG's News}} & BERT-Attack & 74.6 & 15.6 & 0.52 \\
&  \method-mlm  & 76.2 & 15.0 & 0.51 \\
\midrule
\multirow{2}{*}{\textbf{MR}} &BERT-Attack& 83.2&12.8&0.52 \\
&  \method-mlm & 84.3 & 11.5 & 0.53 \\
\bottomrule
\end{tabular}
}
\caption{Attack results of different substitution strategies, where \method-mlm is replaced with the same strategy of word finder as BERT-Attack.} 
\label{tabel:generalization_all}
\end{table}

\end{document}